# HYPERSPECTRAL IMAGING-BASED PERCEPTION IN AUTONOMOUS DRIVING SCENARIOS: BENCHMARKING BASELINE SEMANTIC SEGMENTATION MODELS

*Imad Ali Shah[1,c], Jiarong Li[1], Martin Glavin[1], Edward Jones[1], Enda Ward[2], Brian Deegan[1]*

[1]School of Engineering, University of Galway, University Road, Galway, Ireland
[2]Valeo Vision Systems, Tuam, Co. Galway, Ireland

## ABSTRACT

Hyperspectral Imaging (HSI), known for its advantages over traditional RGB imaging in remote sensing, agriculture, and medicine, has recently gained attention for enhancing Advanced Driving Assistance Systems (ADAS) perception. A few HSI datasets such as HyKo, HSI-Drive, HSI-Road, and Hyperspectral City have been made available. However, a comprehensive evaluation of semantic segmentation models (SSM) using these datasets is lacking. To address this gap, we evaluated the available annotated HSI datasets on four deep learning-based baseline SSMs i.e. DeepLab v3+, HRNet, PSPNet, and U-Net along with its two variants: Coordinate Attention (UNet-CA) and Convolutional Block-Attention Module (UNet-CBAM). The original models' architectures were adapted to handle the varying spatial and spectral dimensions of the datasets. These baseline SSMs were trained using class-weighted loss function for individual HSI datasets and were evaluated over mean-based metrics i.e. intersection over union (IoU), recall, precision, F1 score, specificity, and accuracy. Our results indicate that UNet-CBAM which extracts channel-wise feature extraction, outperforms other SSMs and shows the potential to leverage spectral information for enhanced semantic segmentation. This study establishes a baseline SSM-based benchmark on available annotated datasets for future evaluation of HSI-based ADAS perception. However, the limitations of current HSI datasets, such as limited dataset size, high class imbalance, and lack of fine-grained annotations, remain significant constraints for developing robust SSMs for ADAS applications.

## 1. INTRODUCTION

Semantic segmentation plays a critical role in Advanced Driver Assistance Systems (ADAS) by enabling detailed scene understanding and object identification through per-pixel classification [1]. While ADAS predominantly relies on traditional RGB imaging, Hyperspectral Imaging (HSI) presents considerable advantages. HSI can capture hundreds of intensities across narrow spectral bands, including wavelengths beyond human visibility. This spectrally dense information enables more accurate analysis of material composition and object classification, making HSI a powerful tool, that has been proven in diverse fields such as ecosystem monitoring and agriculture [2] as well as medicine [3].

Recent advances in HSI sensor technology, particularly the development of snapshot hyperspectral sensors, have made these cameras smaller, cheaper, and capable of real-time video capturing [4]. This progress opens new possibilities for HSI in dynamic and real-time applications such as ADAS and autonomous driving (ADAS/AD). HSI has the potential to address key limitations of RGB imaging, such as metamerism [5], and can thus improve object identification, tracking, and general scene understanding in ADAS/AD.

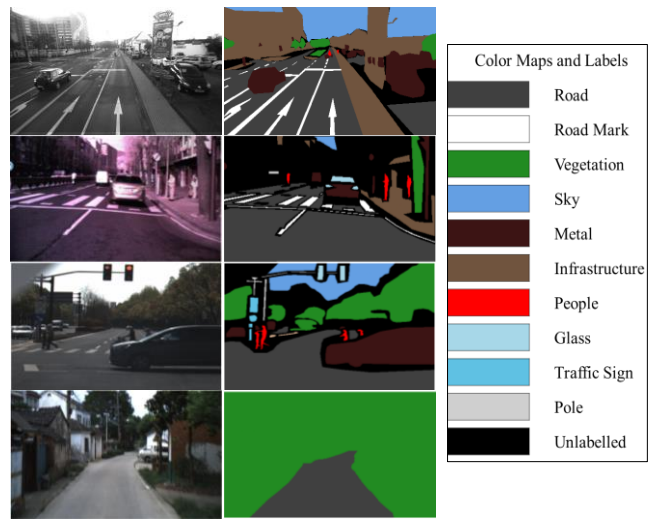

Fig. 1. Available annotated hyperspectral imaging (HSI) datasets: Sample (left) RGB image and (right) ground truth labels from (row-1) Hyko2-VIS, (row-2) HSI-Drive v2, (row-3) Hyperspectral City (HS-City), and (row-4) HSI-Road respectively.

Despite the advantages, HSI faces unique challenges in ADAS/AD scenarios as compared to other fields, such as varying light, weather, and fast-moving objects. The latter is particularly of concern due to HSI sensors' longer acquisition image. These constraints have resulted in a limited number of annotated HSI datasets, such as HyKo [6], HSI-Drive [7][8], Hyperspectral City (HS-City) [9], and HSI-Road [10], sample images shown in Fig 1. Though valuable, these datasets are generally small and with limited diversity of driving conditions. Moreover, the high-dimensional nature of HSI data has also restricted their evaluation of well-generalized semantic segmentation models (SSM) and are primarily evaluated individually. This raises a need to develop baseline SSMs that can provide a foundation for comparing model performance in HSI-based semantic segmentation for ADAS/AD applications.

To establish baseline SSM for HSI in ADAS/AD, this work evaluated six architectures across all four annotated HSI datasets. The models include DeepLab v3+ [11], HRNet [12], PSPNet [13], and U-Net [14] with two variants i.e. U-Net with Coordinate Attention (UNet-CA) [15], and U-Net with Convolutional (Conv) Block-Attention Module (UNet-CBAM) [16]. To accommodate the varying spatial and spectral dimensions of the HSI datasets, these models were adapted to handle input data of different sizes and spectral resolutions. Our main contributions are as follows:

- Evaluation of publicly available annotated HSI datasets for ADAS/AD scenarios.
- Evaluation of four baselines and two U-Net variants for HSI-based semantic segmentation in ADAS/AD.

The remaining paper is structured as follows: Section 2 reviews related work, Section 3 describes the processing of the HSI datasets and experimental setup, Section 4 presents experimental results, and Section 5 concludes the paper.

## 2. RELATED WORK

The use of deep learning techniques for HSI segmentation in ADAS/AD is limited, despite their extensive use in domains like remote sensing. CNN has demonstrated superior performance in extracting HSI spectral-spatial features by utilizing 1D, 2D, and 3D Convs. Others including RNNs, GANs, DBNs, GCNs, and Transformer-based ViT have also demonstrated potential for HSI [17]. However, their application to ADAS/AD-based HSI has yet to be explored.

Most research in ADAS/AD perception has focused on RGB and multi-modal imaging. Benchmark datasets like KITTI [18] and Cityscapes [19] for RGB, while nuScenes [20], Waymo Open Dataset [21], KAIST Multi-Spectral [22], and FLIR ADAS Thermal [23] are widely considered standards for multimodal based segmentation.

HSI-based datasets are limited in number but are recently gaining attention. As shown in Table 1, the available fully annotated datasets such as HyKo v1-v2, HSI-Drive v1-v2, HS-City v1-v2, and HSI-Road. While these datasets provide a foundation, their dataset size and diversity remain insufficient for robust ADAS/AD applications.

Currently, there are no standardized baseline SSMs for HSI in ADAS/AD, as researchers are primarily evaluating their datasets due to high-dimensional HSI data. This lack of standardized benchmarking affects the comparative analysis in understanding the potential of HSI for ADAS/AD. Our work addresses this gap by establishing baseline SSM benchmarks across available datasets. This effort not only facilitates comparative analysis but also provides the foundation for future work in HSI-based applications for ADAS/AD.

TABLE I. DETAILS OF HSI DATASETS FOR ADAS/AD SCENARIOS

| Datasets | Spatial Dimension | Bands | Classes | Range (nm) | No of Images |
|---|---|---|---|---|---|
| HyKo2-VIS | 254x512 | 15 | 10 | 470-630 | 163 |
| HyKo2-NIR | 214x407 | 25 | 10 | 630-975 | 78 |
| HSI-Drive v2 | 209x416 | 25 | 9 | 600-975 | 752 |
| HS-City v2 | 1422x1889 | 128 | 19 | 450-950 | 1,330 |
| HSI-Road | 384x192 | 25 | 2 | 680-960 | 3,799 |

TABLE II. RE-LABELED CLASSES BASED PIXELS DISTRIBUTION (IN MILLIONS) ACROSS HSI DATASETS (NO RELABELLING FOR HSI-ROAD)

| Class Labels | HyKo v2 | | HSI-Drive v2 | | HS-City v2 | |
|---|---|---|---|---|---|---|
| | Count | % | Count | % | Count | % |
| Road | 11.73 | 47.86 | 26.58 | 61.11 | 154.49 | 33.87 |
| Vegetation | 6.06 | 24.71 | 9.3 | 21.38 | 2.00 | 0.44 |
| Sky | 2.73 | 11.13 | 2.51 | 5.76 | 0.901 | 0.20 |
| Metal | 0.861 | 3.51 | 1.29 | 2.97 | 153.16 | 33.58 |
| Infrastructure | 2.81 | 11.45 | 2.29 | 5.27 | 143.86 | 31.54 |
| People | 0.058 | 0.02 | 0.21 | 0.48 | 1.7 | 0.37 |
| Road Marking | 0.32 | 1.32 | 1.32 | 3.04 | - | - |
| Glass | - | - | 0.245 | 0.56 | - | - |
| Unlabeled* | 3.27 | - | 22.34 | - | 234.24 | - |

*Ignored in model training and not counted in above percentage calculation*

## 3. METHODOLOGY AND EXPERIMENTATION

### 3.1. Datasets and Pre-Processing

In this work, we utilized the latest version of the four publicly available and annotated HSI datasets: HyKo v2, HSI-Drive v2, HS-City v2, and HSI-Road. These datasets vary significantly in spatial-spectral resolution, and the number of labeled classes, as illustrated in Table 1.

Current HSI datasets do not follow a standardized format, with imbalanced classes, inconsistent annotation, and varying spectral-spatial resolution. For instance, HSI-Road has only two classes (Road and Others) whereas HS-City v2 has 19 classes, making direct comparisons difficult. To address these issues, we redefined common class labels in the existing annotations of these datasets. These consolidated labels include Road, Vegetation, Sky, Metal (Cars, Traffic Signs, Poles, fences, etc.), Infrastructure (buildings, sidewalks, etc.), and People. While HSI-Road does not align with this structure due to its two-class format, it was retained in the analysis for comprehensive SSM evaluation. Moreover, Road Markings and Glass labels were retained in HSI-Drive, due to potential relevance for material analysis in ADAS/AD applications. The redefined labels and their pixel-wise distribution are listed in Table 2. Other than subsampling of HS-City v2 spatial dimension to 355x472 to expedite model training, given its large size, no other preprocessing or data augmentation was performed. As focus was to keep the dataset's integrity for efficient and comparable baseline SSMs.

### 3.2. Experimentation Setup

#### 3.2.1. Model Selection

Four baseline SSM models were evaluated: U-Net, DeepLab v3+, PSPNet, and HRNet. These models were chosen based on their proven effectiveness in handling the complexity of HSI data. In addition, two variants of U-Net were also included i.e. Coordinate Attention (UNet-CA), and Conv Block-Attention Module (UNet-CBAM). No pre-trained weights or backbones were used in model training. A brief overview of these SSMs is as follows:

- **DeepLab v3+** [11]: From the family of DeepLab, v3+ leverages Atrous Conv and Spatial Pyramid-based pooling for multi-scale feature extraction and object localization.
- **HRNet** [12]: HRNet maintains high-resolution feature representations throughout the segmentation process and preserves fine details.
- **PSPNet** [13]: PSPNet is based on Pyramid based Pooling module, which facilitates the extraction of multi-scale contextual information.
- **U-Net** [14]: U-Net is a standard encoder-decoder-based CNN and is widely used in segmentation tasks for its ability to preserve fine spatial information through skip connections.
- **UNet-CA** [15]**:** A variant of U-Net, incorporates coordinate attention mechanisms (AM) to enhance the model's focus on important spatial and spectral regions.
- **UNet-CBAM** [16]: Another U-Net variant, that integrates CBAM to improve feature selection through channel and spatial wise AM. This channel-wise feature extraction in CBAM is crucial for HSI to leverage the spectrally dense information.

#### 3.2.2. Model Adaptation

To accommodate the varying input spatial and spectral dimensions of the HSI datasets, we modified the baseline SSMs. The modifications are as follows:

- **Input and Output Layers Modification**: The input layer of models was adjusted to individual HSI dataset's spectral and spatial dimensions, and the output layers to the required number of predicted classes.
- **Intra-Modules Layers**: The non-standardized spatial dimensions of HSI datasets required careful adjustment of intermediate layers. For instance, the U-Net model was adjusted to handle the dynamic padding for adopting dimension mismatches between encoder and decoder blocks. These modifications ensured model integrity and feature integration with smooth information flow.

TABLE III. OVERVIEW OF THE HSI-DATASETS, MODEL HYPERPARAMETERS AND EXPERIMENTAL SETTINGS

| Detail | HyKo v2 | | HSI-Drive v2 | HS-City v2 | HSI-Road |
|---|---|---|---|---|---|
| | VIS | NIR | | | |
| Batch Size | 16 | 8 | 16 | 8 | 16 |
| AS* | 2 | 2 | 2 | 4 | 2 |
| Optimizer | AdaBelief [24] ($\beta_1$: 0.9 and $\beta_2$: 0.99) | | | | |
| Scheduler | ReduceOnPlateau (patience: 4, factor: 0.9, min: 5e-7) | | | | |
| LR* | 0.0006 with a restart to 90% using scheduler | | | | |
| Loss | Dice-coefficient and Class-weighted Cross-entropy | | | | |
| Activation | LeakyReLU | | | | |
| Hardware | Core i9-13900K, 64GB RAM, and Nvidia RTX 4090 TI | | | | |

*AS: Accumulation Step, and LR: Learning Rate

#### 3.2.3. Setup

Table 3 illustrates a comprehensive detail of the setup and hyperparameters for the model training. These parameters were selected based on optimal performance using Grid Search. All models were trained for 300 epochs using the class weighted-based Cross-Entropy loss function. Learning rate (lr) was scheduled using ReduceOnPlateau with a restart to 90% lr upon reaching the minimum threshold. This approach helped in escaping local minima and promoting exploration of the parameter space, thus improving convergence in challenging hyperspectral optimization tasks.

To comprehensively evaluate and gain insights into model performance, we employed Intersection over Union (IoU), Precision (Prec), Recall (Rec), F1, Specificity (Spec), and Accuracy (Acc) metrics. These metrics were averaged i.e. mean (m) across all classes to provide a holistic models' assessment. Moreover, models were trained using mixed precision, with no regularization or early stopping.

### 4. RESULTS AND DISCUSSION

As shown in Table 4, U-Net and its variants, especially UNet-CBAM, demonstrated the most promising results:

- UNet-CBAM consistently outperformed the other models, achieving mIoU and mF1 values of 65.31% and 73.45% for HyKo2-VIS, 86.56% and 92.80% for HSI-Drive v2, 87.23% and 92.06% for HS-City v2, and 96.56% and 98.26% for HSI-Road, respectively.
- The integration of attention mechanisms (AM), particularly the channel-wise extraction in CBAM, significantly enhanced feature extraction. CBAM's ability to capture both channel (spectral) and spatial correlations within HSI proved beneficial, as illustrated in Fig 2, CBAM can be seen to preserve fine details better than the other models.

However, the performance of these models is constrained by inherent limitations of the HSI datasets, such as a limited number of images as shown in Table 2, and highly imbalanced classes with coarse labeling. As shown in Fig 2, the example of HyKo2-VIS (mid column) lacks Road Mark annotation, and the models try to segment it based on their previously learned features during other training samples. Similarly, in HSI-Drive v2 (left column) the most complex region of the scene is not annotated. These limitations directly affect the robustness and generalizability of SSMs across diverse ADAS/AD conditions.

TABLE IV.  EXPERIMENTATION RESULTS OF HSI-DATASETS OVER BASELINE SEMANTIC SEGMENTATION MODELS

| Dataset | Model | Metrics (mean over all classes) | | | | | |
|---|---|---|---|---|---|---|---|
| | | IoU | Prec | Rec | F1 | Spec | Acc |
| HyKo2-VIS | DeepLabv3+ | 63.20 | 70.89 | 72.51 | 71.68 | 98.34 | 97.65 |
| | HRNet | 61.25 | 70.05 | 71.26 | 70.63 | 98.19 | 97.44 |
| | PSPNet | 42.04 | 49.60 | 53.76 | 51.54 | 96.43 | 95.16 |
| | U-Net | 61.78 | 70.34 | 71.46 | 70.86 | 98.25 | 97.49 |
| | UNet-CA | 64.15 | 71.70 | 73.13 | 72.37 | **98.58** | 97.97 |
| | UNet-CBAM | **65.31** | **73.77** | **73.15** | **73.45** | **98.58** | **97.98** |
| HyKo2-NIR | DeepLabv3+ | **80.74** | **85.42** | **94.59** | **89.77** | 99.30 | 99.36 |
| | HRNet | 75.97 | 79.52 | 92.69 | 85.60 | 98.95 | 99.06 |
| | PSPNet | 71.94 | 76.09 | 88.87 | 81.98 | 98.65 | 98.93 |
| | U-Net | 72.79 | 76.92 | 75.34 | 76.12 | 99.44 | 99.51 |
| | UNet-CA | 74.14 | 76.57 | 77.13 | 76.85 | 99.44 | 99.52 |
| | UNet-CBAM | 75.38 | 77.80 | 77.37 | 77.59 | **99.58** | **99.65** |
| HSI-Drive v2 | DeepLabv3+ | 79.94 | 86.90 | 88.11 | 87.39 | 99.45 | 99.26 |
| | HRNet | 76.03 | 85.05 | 84.89 | 84.89 | 99.12 | 98.84 |
| | PSPNet | 55.27 | 63.90 | 67.81 | 65.74 | 97.91 | 97.51 |
| | U-Net | 84.07 | 88.83 | 91.44 | 90.10 | 99.59 | 99.42 |
| | UNet-CA | 86.34 | 91.32 | **93.07** | **92.17** | 99.64 | 99.51 |
| | UNet-CBAM | **86.56** | **91.34** | 92.80 | 92.06 | **99.70** | **99.55** |
| HS-City v2 | DeepLabv3+ | 85.12 | 89.37 | 91.52 | 90.36 | 99.36 | 99.09 |
| | HRNet | 78.23 | 85.99 | 86.00 | 85.77 | 98.18 | 97.42 |
| | PSPNet | 70.72 | 77.82 | 78.85 | 78.23 | 98.29 | 97.52 |
| | U-Net | 86.08 | 90.54 | 92.22 | 91.27 | 99.41 | 99.15 |
| | UNet-CA | 86.91 | 91.04 | 92.50 | 91.69 | 99.47 | 99.22 |
| | UNet-CBAM | **87.23** | **91.22** | **93.13** | **92.06** | **99.49** | **99.28** |
| HSI-Road | DeepLabv3+ | 96.15 | 98.12 | 97.92 | 98.02 | 98.12 | 98.53 |
| | HRNet | 95.62 | 97.92 | 97.58 | 97.75 | 97.92 | 98.32 |
| | PSPNet | 93.24 | 96.71 | 96.19 | 96.44 | 96.71 | 97.39 |
| | U-Net | 96.44 | 98.33 | 98.02 | 98.17 | 98.30 | 98.65 |
| | UNet-CA | 96.55 | 98.28 | 98.18 | 98.23 | 98.28 | 98.69 |
| | UNet-CBAM | **96.56** | **98.34** | **98.19** | **98.26** | **98.33** | **98.70** |

The results presented in this paper establish a standardized baseline SSMs-based benchmark, and the results of AM-based models provide potential for HSI-based segmentation in ADAS/AD applications. However, addressing the highlighted limitations in HSI datasets, such as limited dataset size and diversity, coarse annotations, etc., in future works will be the key to more robust HSI-based solutions.

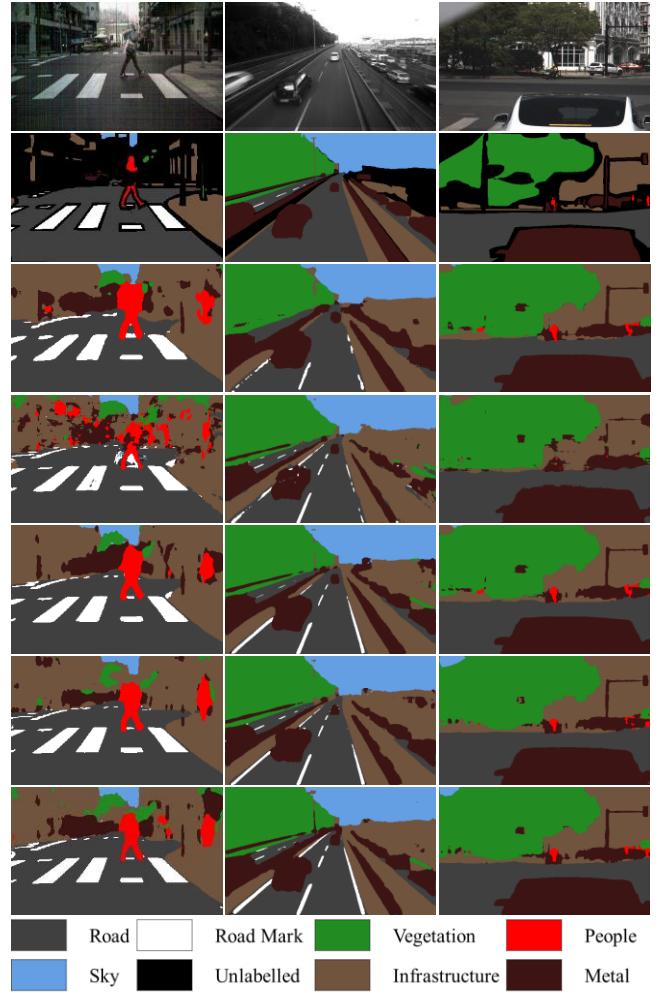

Fig. 2.  Segmentation results of (left) HSI-Drive v2, (mid) HyKo2-VIS, and (right) HS-City v2 datasets with sample images in row-1 and true labels in row-2. Whereas rows 3-7 present the respective segmentations by DeepLabv3+, HRNet, U-Net, UNet-CA, and UNet-CBAM.

## 5. CONCLUSION

This paper presents a comprehensive evaluation of baseline semantic segmentation models (SSM) for hyperspectral imaging (HSI) datasets for Advanced Driver Assistance Systems and Autonomous Driving (ADAS/AD). It assesses U-Net, DeepLabv3+, PSPNet, HRNet and two U-Net variants (UNet-CA, UNet-CBAM). Among the evaluated SSMs, UNet-CBAM outperformed other models, providing the potential of channel-wise attention mechanisms (AM) for effectively leveraging spectral-dense HSI datasets. However, the inherent limitations of available datasets, including small dataset sizes, non-standard spectral-spatial dimensions, high class imbalance, and coarse annotations, remain significant challenges. Future research should address these limitations by developing larger, more diverse, and fine-grained annotation-based datasets for developing more robust and generalizable SSMs. The results presented in this paper provide a foundation for the further development and evaluation of HSI-based segmentation models in the ADAS/AD domain.

**Acknowledgment:** This work was supported, in part, by Science Foundation Ireland grant 13/RC/2094_P2, and Valeo Vision Systems.